\documentclass[sigconf, screen]{acmart}
\usepackage{booktabs} 
\usepackage{float}
\usepackage{algorithm}
\usepackage{algpseudocode}

\usepackage{amssymb}
\usepackage{subcaption}
\usepackage{caption}
\usepackage{pifont}
\usepackage{tabularx}
\usepackage{pifont} 
\newcommand{\cmark}{\ding{51}} 
\newcommand{\xmark}{\ding{55}} 

\setcopyright{none}
\renewcommand\footnotetextcopyrightpermission[1]{}
\AtBeginDocument{%
  }

\begin{document}

\title{Towards Graph-Based Privacy-Preserving Federated Learning: ModelNet - A ResNet-based Model Classification Dataset}

\author{Abhisek Ray}
 \email{ar@ece.au.dk}
 \orcid{0000-0002-0551-5674}
\author{Lukas Esterle}
\email{lukas.esterle@ece.au.dk}
\orcid{0000-0002-0248-1552}
\affiliation{%
  \institution{Aarhus University}
  \city{Aarhus}
  \country{Denmark}
}

\renewcommand{\shortauthors}{Ray and Esterle}

\begin{abstract}
    Federated Learning (FL) has emerged as a powerful paradigm for training machine learning models across distributed data sources while preserving data locality. However, the privacy of local data is always a pivotal concern and has received a lot of attention in recent research on the FL regime. Moreover, the lack of domain heterogeneity and client-specific segregation in the benchmarks remains a critical bottleneck for rigorous evaluation. In this paper, we introduce ModelNet, a novel image classification dataset constructed from the embeddings extracted from a pre-trained ResNet50 model. First, we modify the CIFAR100 dataset into three client-specific variants, considering three domain heterogeneities (homogeneous, heterogeneous, and random). Subsequently, we train each client-specific subset of all three variants on the pre-trained ResNet50 model to save model parameters. In addition to multi-domain image data, we propose a new hypothesis to define the FL algorithm that can access the anonymized model parameters to preserve the local privacy in a more effective manner compared to existing ones. ModelNet is designed to simulate realistic FL settings by incorporating non-IID data distributions and client diversity design principles in the mainframe for both conventional and futuristic graph-driven FL algorithms. The three variants are ModelNet-S, ModelNet-D, and ModelNet-R, which are based on homogeneous, heterogeneous, and random data settings, respectively. To the best of our knowledge, we are the first to propose a cross-environment client-specific FL dataset along with the graph-based variant. Extensive experiments based on domain shifts and aggregation strategies show the effectiveness of the above variants, making it a practical benchmark for classical and graph-based FL research. The dataset and related code are available \href{https://github.com/rayabhisek123/ModelNet}{here}\footnote{https://github.com/rayabhisek123/ModelNet}.
\end{abstract}





\maketitle
\pagestyle{plain}

\section{Introduction}
Federated learning (FL) is a decentralized machine learning paradigm enabling multiple devices to collaboratively train a model without sharing raw data, thus preserving privacy \cite{mcmahan2017communication}. 
The diversity of environmental settings, such as data distribution and relationships between edge models, plays a significant role in convergence, generalization, and fairness. 
The homogeneous, heterogeneous, or random nature of these local data, i.e., non-Independent and Identically Distributed (non-IID), can lead to model drift, hence affecting communication and model aggregation strategies. Furthermore, the nature of inter-client relationships influences collaborative dynamics and model alignment. Hence, modeling and leveraging these factors are crucial for robust and efficient FL systems, especially in real-world deployments where uniformity across clients cannot be assumed. Various approaches have tackled the problem of handling non-IID data in clustered FL~\cite{ghosh2020efficient, briggs2020federated, duan2021flexible, domini2024proximity, domini2024field}.

\begin{table*}
	\begin{center}
		\caption{{Comparison of various FL algorithms with ModelNet}}
		\tabcolsep=0.13cm
		\scalebox{0.885}{
			    \begin{tabular}{lccccc}
                    \hline
                    \textbf{Partitioning Method} & \textbf{Data Balance} & \textbf{Class Distribution Control} & \textbf{Semantic Grouping} & \textbf{Heterogeneity Modeling} & \textbf{Realism for FL} \\
                    \hline
                    IID Partitioning \cite{mcmahan2017communication} & High & \xmark & \xmark & \xmark & \xmark \\
                    Distribution Partitioning \cite{mcmahan2017communication} & Moderate & \cmark (Manual) & \xmark & \cmark & \cmark \\
                    Dirichlet Partitioning \cite{wang2020federated} & Moderate & \cmark (Tunable $\alpha$) & \xmark & \xmark & \cmark \\
                    InnerDirichlet Partitioning \cite{acar2021federated} & Moderate & \cmark (Class-wise) & \xmark & \cmark & \cmark \\
                    Linear Partitioner \cite{flowerDatasets} & High & \xmark & \xmark & \xmark & \xmark \\
                    Square Partitioner \cite{flowerDatasets} & High & \xmark & \xmark & \xmark & \xmark \\
                    Proximity-based Partitioning \cite{domini2025profed} & Moderate–High & \cmark (Regional Skew Control) & \cmark (Geographical) & \cmark (Clustered Non-IID) & \cmark  \\
                    \textbf{ModelNet-R Algorithm (Ours)} & High & \cmark (Uniform Subsets) & \xmark (Random) & \cmark (Mild) & \cmark \\
                    \textbf{ModelNet-D Algorithm (Ours)} & High & \cmark (Clustering) & \cmark (Diverse) & \cmark & \cmark \\
                    \textbf{ModelNet-S Algorithm (Ours)} & High & \cmark (Clustering) & \cmark (Similar) & \cmark & \cmark \\
                    \hline
                    \end{tabular}}
		\label{table:algorithms}
	\end{center}
\end{table*}

\begin{table*}[t]
  \caption{{Comparison of existing FL image classification datasets with our proposed ModelNet variants}}
  \label{tab:dataset_comparison}
  \scalebox{0.835}{
  \begin{tabular}{@{}lcccccccc@{}}
    \toprule
    \textbf{Feature / Dataset} & \textbf{EMNIST}\cite{cohen2017emnist} & \textbf{CIFAR-10}\cite{krizhevsky2009cifar} & \textbf{CIFAR-100}\cite{krizhevsky2009cifar} & \textbf{TinyImageNet}\cite{le2015tiny} & \textbf{ProFed} \cite{domini2025profed} & \textbf{ModelNet-R} & \textbf{ModelNet-S} & \textbf{ModelNet-D} \\
    \midrule
    Natural Client Split        & \cmark & \xmark & \xmark & \xmark & Synthetic & Synthetic & Synthetic & Synthetic \\
    Class Count                 & 62     & 10     & 100    & 200  & Varies (10–100)  & 15/subset & 15/subset & 15/subset \\
    Semantic Control            & \xmark & \xmark & \xmark & \xmark & \cmark & Random    & \cmark    & \cmark    \\
    Dataset Size                & Medium & Small  & Medium & Medium-Large & Medium–Large & Large & Large & Large \\
    Non-IID Evaluation          & \cmark & \cmark & \cmark & \xmark & \cmark  & \cmark  & \cmark    & \cmark    \\
    Use in FL Research          & Established & Common & Common & Emerging & New & New & New & New \\
    Personalization Suitability & Limited & Limited & Limited & Limited & High & High & High & High \\
    \bottomrule
  \end{tabular}}
\end{table*}
\begin{figure*}[t!]
    \centering
    \captionsetup{justification=centering}
    \includegraphics[width=0.975\linewidth]{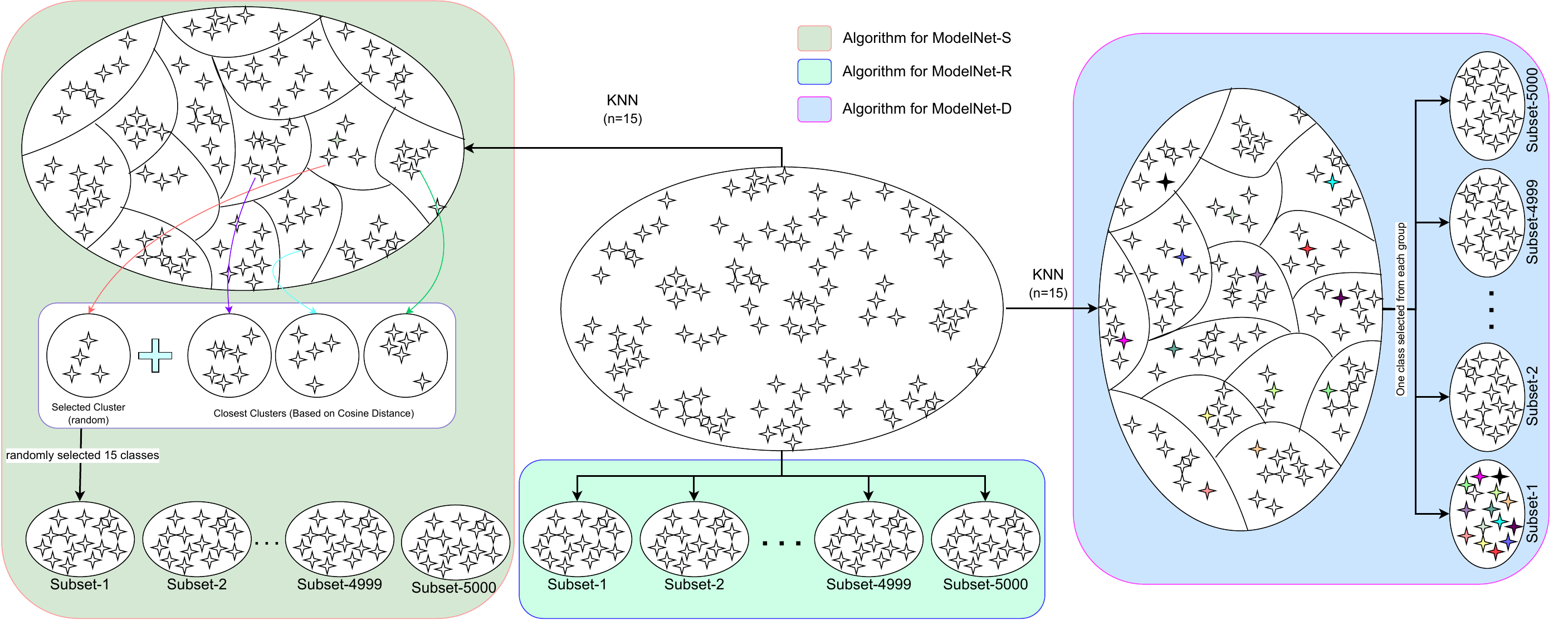}
    \caption{{Overall architecture for three variants of ModelNet (ModelNet-R, ModelNet-D, and ModelNet-S)}}
    \label{fig:Architecture}
\end{figure*}

Existing image classification datasets in federated learning (FL), such as EMNIST~\cite{cohen2017emnist}, CIFAR-10/100~\cite{krizhevsky2009cifar}, and Fashion-MNIST~\cite{xiao2017fashionmnist}, offer a solid foundation for studying non-IID learning scenarios. However, they often fall short in terms of controlling semantic similarity between classes across clients, limiting their ability to explore how intra-subset semantic structure affects federated performance. For example, data sets such as EMNIST or CIFAR-10 have limited class diversity and fixed label sets, leaving little room to evaluate the impact of class similarity or diversity within and across clients. To overcome these limitations, we introduce ModelNet, a new dataset derived from CIFAR-100 and organized into three distinct variants: ModelNet-R (random class selection), ModelNet-S (semantically similar classes), and ModelNet-D (semantically diverse classes). Each variant is large-scale and facilitates fine-grained control over the statistical and semantic heterogeneity among clients. This design enables comprehensive benchmarking of FL algorithms in varied interclient relationships. It supports the study of model personalization, generalization, and robustness to semantic skew, which existing benchmarks do not fully address. Tables \ref{table:algorithms} and \ref{tab:dataset_comparison} compare the proposed ModelNet dataset and its underlying algorithm with other conventional datasets and algorithms in the FL regime, respectively. The algorithm behind the ProFed benchmark \cite{domini2025profed} is very competitive, but it only controls inter-client relations in terms of skewness. However, the ModelNet algorithm can govern both inter- and intra-client affinity. Like ProFed, ModelNet can be scaled to any other dataset beyond CIFAR-100.

All of the above benchmarks explain various client affinities in terms of data interpretability. In addition to data interpretation, our dataset lays the foundation for graph-based model interpretation towards privacy-preserving FL. The graph-based dataset variant can also be expanded to any other client-oriented base model beyond pre-trained ResNet50. The contribution of the paper can be summarized as follows.
\begin{itemize}
    \item We propose \textbf{ModelNet}, a large-scale image classification benchmark tailored for federated learning, derived from CIFAR-100 and designed to systematically control semantic and statistical heterogeneity.
    
    \item We construct three distinct variants of ModelNet—\textbf{ModelNet-R}, \textbf{ModelNet-D}, and \textbf{ModelNet-S} using randomized, semantically diverse, and semantically similar sampling strategies, respectively. Each variant contains 5000 subsets to simulate federated clients. We also introduce a parameter-based variant that lays the foundation for graph-based FL interpretation. 
    
    \item We extensively evaluate all three variants on multiple metrics for more fine-grained analysis and benchmarking compared to existing FL datasets, which proves its effectiveness in multi-domain FL settings.
\end{itemize}




\section{ModelNet}
\label{sec:ModelNet}
The ModelNet dataset serves as a large-scale benchmark for federated learning by simulating non-IID conditions through artificial partitioning of CIFAR-100 classes between FL clients. To address key challenges such as model personalization, generalization under client diversity, and robustness to semantic skew, we introduce three variants of ModelNet, each constructed using a different algorithm, as illustrated in Fig. \ref{fig:Architecture}. We provide an overview of these three variants along with their corresponding algorithms below:

    \paragraph{\textbf{ModelNet-R}}
        To create a diverse training setup for class-subset evaluation, we randomly sample $N = 5000$ subsets from a fixed pool of $100$ semantic classes, as shown in Alg. \ref{alg:ModelNet-R}. Each subset $\mathcal{S}_i \subset \mathcal{C}$, where $\mathcal{C} = \{0, 1, \ldots, 99\}$, contains exactly $15$ distinct classes sampled uniformly without replacement. That is, for each $i \in \{1, \ldots, 5000\}$, we draw
        \begin{equation}
            \mathcal{S}_i \sim \text{UniformSample}(\mathcal{C},\ 15\ \text{classes}).
        \end{equation}
        
        No constraints are imposed on the overlap or similarity between subsets, so repetitions and partial intersections between subsets are possible and expected. The resulting collection $\{\mathcal{S}_i\}_{i=1}^{5000}$ can be used for downstream training, evaluation, or diversity analysis tasks. An optional random seed ensures reproducibility of the generated subset configuration.
        {
        \begin{algorithm}
        \caption{Generation of ModelNet-R}
        \label{alg:ModelNet-R}
        \begin{algorithmic}[1]
        \Require Dataset $\mathcal{D}$ with class folders $\mathcal{C}$, parameters: $N$ (subsets), $S$ (classes per subset), $I$ (images per class, optional), $seed$, $copyMethod$
        \Ensure Output subsets $\{\mathcal{S}_i\}_{i=1}^N$
        \State Init random seed if provided
        \State Sort class dirs: $\mathcal{C} = \{c_1, \ldots, c_{|\mathcal{C}|}\}$
        \If{$|\mathcal{C}| < S$} \textbf{throw} error \EndIf
        \For{$i = 1$ to $N$}
            \State Sample $\mathcal{S}_i \subset \mathcal{C}, |\mathcal{S}_i| = S$
            \ForAll{$c \in \mathcal{S}_i$}
                \State List images $\mathcal{I}_c$
                \If{$I$ specified}
                    \If{$|\mathcal{I}_c| < I$} \textbf{throw} error \EndIf
                    \State Sample $\mathcal{I}_c' \subseteq \mathcal{I}_c, |\mathcal{I}_c'| = I$
                \Else
                    \State $\mathcal{I}_c' \gets \mathcal{I}_c$
                \EndIf
                \State Copy/move/symlink $\mathcal{I}_c'$ to subset dir
            \EndFor
        \EndFor
        \State \Return $\{\mathcal{S}_i\}$
        \end{algorithmic}
        \end{algorithm}}
        
    \paragraph{\textbf{ModelNet-D}}
        {
        \begin{algorithm}
        \caption{Algorithm for the generation of ModelNet-D}
        \label{alg:ModelNet-D}
        \begin{algorithmic}[1]
        \Require Dataset $\mathcal{D} = \{(c_i, I_{c_i})\}$ with classes $c_i$, pretrained model $\mathcal{M}$, 
        parameters $N$ (subsets), $K$ (clusters), $I$ (images/class), seed, copy method
        \Ensure Subsets $\{\mathcal{S}_i\}_{i=1}^N$ with dissimilar classes
    
        \Comment{\color{gray}Step 1: Extract class embeddings\color{black}}
    
        \ForAll{$c \in \mathcal{C}$}
          \State $\mathbf{e}_c = \frac{1}{M} \sum_{m=1}^{M} \mathcal{M}(I_c^{(m)})$
        \EndFor
    
        \Comment{\color{gray}Step 2: Cluster classes\color{black}}
        \State $\{\mathcal{C}_k\}_{k=1}^K = \text{KMeans}(\{\mathbf{e}_c\})$
        
        \Comment{\color{gray}Step 3: Generate subsets\color{black}}
        \ForAll{$i \in [1, N]$}
        \State $\mathcal{S}_i = \{c_k : c_k \sim \text{Uniform}(\mathcal{C}_k)\}_{k=1}^K$
        
        \ForAll{$c \in \mathcal{S}_i$}
        \State $\mathcal{I}_c \sim \text{UniformSample}(I_c, I)$
        \State $\text{Copy/move/symlink } \mathcal{I}_c \to \text{subset directory}$
        \EndFor
        \EndFor
        \end{algorithmic}
        \end{algorithm}}
        As depicted in Alg. \ref{alg:ModelNet-D}, we present a method to generate multiple subsets comprising dissimilar classes from a labeled image dataset by leveraging pretrained deep feature embeddings and unsupervised clustering. First, representative feature vectors for each class are obtained by averaging embeddings extracted from a pre-trained convolutional neural network (here ResNet50) across a fixed number of images per class. These class embeddings capture the semantics of each category in a high-dimensional feature space. Next, k-means clustering is applied to group classes into distinct clusters of semantically similar classes. To ensure diversity, each subset is constructed by randomly selecting exactly one class from each cluster, thereby maximizing inter-class dissimilarity within the subset. For each selected class, a fixed number of images are randomly sampled for each class in the subset. This approach enables the creation of a large number of subsets that are both diverse and representative, facilitating robust training and evaluation protocols in classification and related tasks without requiring manual annotation or supervision for class similarity.

    \paragraph{\textbf{ModelNet-S}}
    
        {
        \begin{algorithm}
        \caption{Algorithm for the generation of ModelNet-S}
        \label{alg:ModelNet-S}
        \begin{algorithmic}[1]
        \Require Dataset $\mathcal{D}$ with classes $c \in \mathcal{C}$; pretrained model $M$; parameters $maxImgs$, $K$, $N$, $S$, $I$, $topK$, $copyMethod$, $seed$
        \Ensure Subsets $\{\mathcal{S}_i\}_{i=1}^N$, each with $S$ classes of similar semantics 
        
        \State Initialize device and load model $M$ in evaluation mode
        \State Define preprocessing transform $T$
        
        \Comment{\color{gray}Step 1: Compute class embeddings\color{black}}
        \ForAll{$c \in \mathcal{C}$}
            \State $F_c \gets \frac{1}{|\mathcal{I}_c|} \sum_{I \in \mathcal{I}_c} M(T(I))$ \quad with $|\mathcal{I}_c| \leq maxImgs$
        \EndFor
        
        \Comment{\color{gray}Step 2: Cluster classes by embeddings \color{black}}
        \State Perform $k$-means clustering on $\{F_c\}$ into $K$ clusters $\{\mathcal{C}_k\}_{k=1}^K$
        
        \Comment{\color{gray}Step 3: Compute cluster centroids \color{black}}
        \State $G_k \gets \frac{1}{|\mathcal{C}_k|} \sum_{c \in \mathcal{C}_k} F_c$
        
        \Comment{\color{gray}Step 4: Find top-$topK$ similar clusters for each cluster\color{black}}
        \ForAll{$k=1,\dots,K$}
            \State Compute similarity scores $s_{k,j} = \cos(G_k, G_j)$, $j \neq k$
            \State $N_k \gets$ indices of top-$topK$ clusters by similarity
        \EndFor
        
        \Comment{\color{gray}Step 5: Generate $N$ subsets\color{black}}
        \For{$i=1$ \textbf{to} $N$}
            \State Randomly select base cluster $b \in \{1,\ldots,K\}$
            \State Candidate pool $\mathcal{P}_i \gets \mathcal{C}_b \cup \bigcup_{k \in N_b} \mathcal{C}_k$
            \If{$|\mathcal{P}_i| < S$} \textbf{continue} 
            \EndIf
            \State Sample classes $\mathcal{S}_i \subset \mathcal{P}_i$, $|\mathcal{S}_i| = S$
            \ForAll{$c \in \mathcal{S}_i$}
                \State Sample $I$ images from class $c$: $\mathcal{I}_c' \subseteq \mathcal{I}_c$, $|\mathcal{I}_c'|=I$
                \State Copy/move/symlink images $\mathcal{I}_c'$ to output directory 
            \EndFor
        \EndFor
        
        \end{algorithmic}
        \end{algorithm}
        We extract feature embeddings for each class by averaging CNN (here ResNet50) features from a subset of images. Classes are then grouped into clusters using k-means on these embeddings. For each cluster, we identify the most similar clusters based on the cosine similarity of their centroids. To create each subset, we randomly pick a base cluster and form a pool including that cluster and its similar clusters. From this pool, we sample a fixed number of classes and select images per class to form subsets with semantically related classes. This process, which is depicted in Alg. \ref{alg:ModelNet-S}, is repeated to generate multiple subsets containing mostly similar classes.
    


\section{Evaluation methodology and metrics}
\label{sec:Methodology}
    
    \paragraph{\textbf{Subset Class Embedding Diversity}}
        To quantify the intra-subset diversity of class representations, we compute the average pairwise cosine distance between class embeddings within each subset. This metric, referred to as subset class embedding diversity, captures the semantic spread of classes and is defined as:
        \begin{equation}
            \mathcal{D}(S) = \frac{2}{|S|(|S|-1)} \sum_{i < j} \left(1 - \cos\left(\mathbf{e}_i, \mathbf{e}_j\right)\right),
        \end{equation}
        where $S$ is the set of class embeddings $\{\mathbf{e}_1, \dots, \mathbf{e}_{|S|}\}$, and $\cos(\mathbf{e}_i, \mathbf{e}_j)$ denotes the cosine similarity between embeddings $\mathbf{e}_i$ and $\mathbf{e}_j$. A higher value of $\mathcal{D}(S)$ indicates greater diversity among the classes in the subset.

    \paragraph{\textbf{Jaccard Similarity}}
        To quantitatively assess the degree of class overlap among subsets derived from different sampling strategies, we employ the Jaccard Similarity Index. For any two sets $A$ and $B$, the Jaccard similarity is defined as:
        $J(A, B) = \frac{|A \cap B|}{|A \cup B|}.$
        This metric evaluates the proportion of shared elements between two sets relative to their union, with $J = 1$ indicating complete identity and $J = 0$ representing disjoint sets.

    \paragraph{\textbf{Class Occurrence Histogram}}

        The class occurrence histogram reflects the distribution of class labels in a dataset or subset. Given a dataset $\mathcal{D} = {(\mathbf{x}i, y_i)}{i=1}^N$, where each label $y_i \in {1, \dots, C}$ corresponds to input $\mathbf{x}_i$, the class histogram is defined as the vector $\mathbf{h} = [h_1, h_2, \dots, h_C] \in \mathbb{R}^C$, where each component $h_c$ counts the number of samples in class $c$. This and the entropy of the class distribution can be expressed as:
        \begin{equation}
        \begin{aligned}
        h_c &= \sum_{i=1}^N \mathbb{I}[y_i = c], \quad &&\text{for } c \in {1, \dots, C}, \
        \mathcal{H}(\mathbf{h}) &= - \sum_{c=1}^{C} \frac{h_c}{N} \log \left( \frac{h_c}{N} \right)
        \end{aligned}
        \end{equation}
        
        \noindent where $\mathbb{I}[\cdot]$ is the indicator function returning 1 if the condition is true and 0 otherwise. This histogram provides a discrete distribution over class labels. In a balanced subset, each $h_c \approx N/C$; deviations indicate class imbalance. The entropy $\mathcal{H}(\mathbf{h})$ reaches its maximum when all classes appear equally often.


    \paragraph{\textbf{t-SNE plot}}
        A t-SNE plot is a 2D or 3D visualization generated using the t-distributed Stochastic Neighbor Embedding (t-SNE) algorithm, which projects high-dimensional data into a lower-dimensional space while preserving local neighborhood structure. It is widely used in computer vision to qualitatively assess feature separability, cluster structure, or embedding quality in deep learning models.

    \paragraph{\textbf{Redundancy Metric}}
        We define subset redundancy $R$ as the average number of overlapping classes between all unique pairs of subsets within the same dataset variant. Let $\mathcal{S} = \{S_1, S_2, \dots, S_n\}$ denote a collection of $n$ subsets, where each subset $S_i \subseteq \mathcal{C}$, and $\mathcal{C}$ is the global set of classes. The redundancy between subsets $S_i$ and $S_j$ is computed as: $R_{ij} = |S_i \cap S_j|.$
        The overall redundancy $\bar{R}$ is then given by:
        \begin{equation}
            \bar{R} = \frac{2}{n(n-1)} \sum_{1 \leq i < j \leq n} |S_i \cap S_j|.
        \end{equation}
        This metric captures the extent to which class selections are repeated across different subsets, with higher values indicating increased redundancy and reduced subset uniqueness.

    \paragraph{\textbf{Intra-Subset Variance}}
        To quantify the semantic diversity within a single subset, we define the intra-subset variance as the average pairwise distance between the feature embeddings of all classes in that subset. This measure provides insight into how heterogeneous or homogeneous the selected classes are in terms of their learned representations.
        
        Let $S = \{c_1, c_2, \dots, c_k\}$ be a subset of $k$ classes, and let $\phi(c_i) \in \mathbb{R}^d$ denote the embedding vector of class $c_i$ in a semantic or feature space of dimension $d$. The intra-subset variance $\mathcal{V}_{\text{intra}}(S)$ is defined as:
        \begin{equation}
            \mathcal{V}_{\text{intra}}(S) = \frac{2}{k(k-1)} \sum_{1 \leq i < j \leq k} \|\phi(c_i) - \phi(c_j)\|_2^2.
        \end{equation}
        Here, $\|\cdot\|_2$ denotes the Euclidean norm. This expression computes the mean squared distance between all class pairs within a subset, providing a scalar measure of spread or diversity. High intra-subset variance indicates that the selected classes are semantically distant from one another—favorable for evaluating generalization across diverse categories. Low intra-subset variance suggests the classes are closely related in embedding space, useful for stress-testing models on fine-grained or visually similar categories.

    \paragraph{\textbf{Feature Space Coverage}}
        We also compute the feature space coverage by evaluating the trace of the global covariance matrix of feature embeddings to assess how effectively a given dataset variant spans the underlying representation space. If $\mathbf{X} = \{\mathbf{x}_1, \dots, \mathbf{x}_N\}$ denotes the set of deep features extracted from all samples in a dataset subset, the empirical covariance matrix $\Sigma \in \mathbb{R}^{d \times d}$ is computed as:
        \begin{equation}
            \Sigma = \frac{1}{N} \sum_{i=1}^{N} (\mathbf{x}_i - \boldsymbol{\mu})(\mathbf{x}_i - \boldsymbol{\mu})^\top, \quad \text{where} \quad \boldsymbol{\mu} = \frac{1}{N} \sum_{i=1}^{N} \mathbf{x}_i.
        \end{equation}
        We define feature space coverage as $\text{Tr}(\Sigma)$, where $\text{Tr}(\cdot)$ denotes the matrix trace, capturing the total variance across all feature dimensions.
    


\section{Results and Discussion}
\label{sec:Results}
    
    \paragraph{\textbf{Subset Diversity of ModelNet Variants}}

        \begin{figure}
          \centering
          \captionsetup{justification=centering}
            \centering
            \includegraphics[width=.95\linewidth]{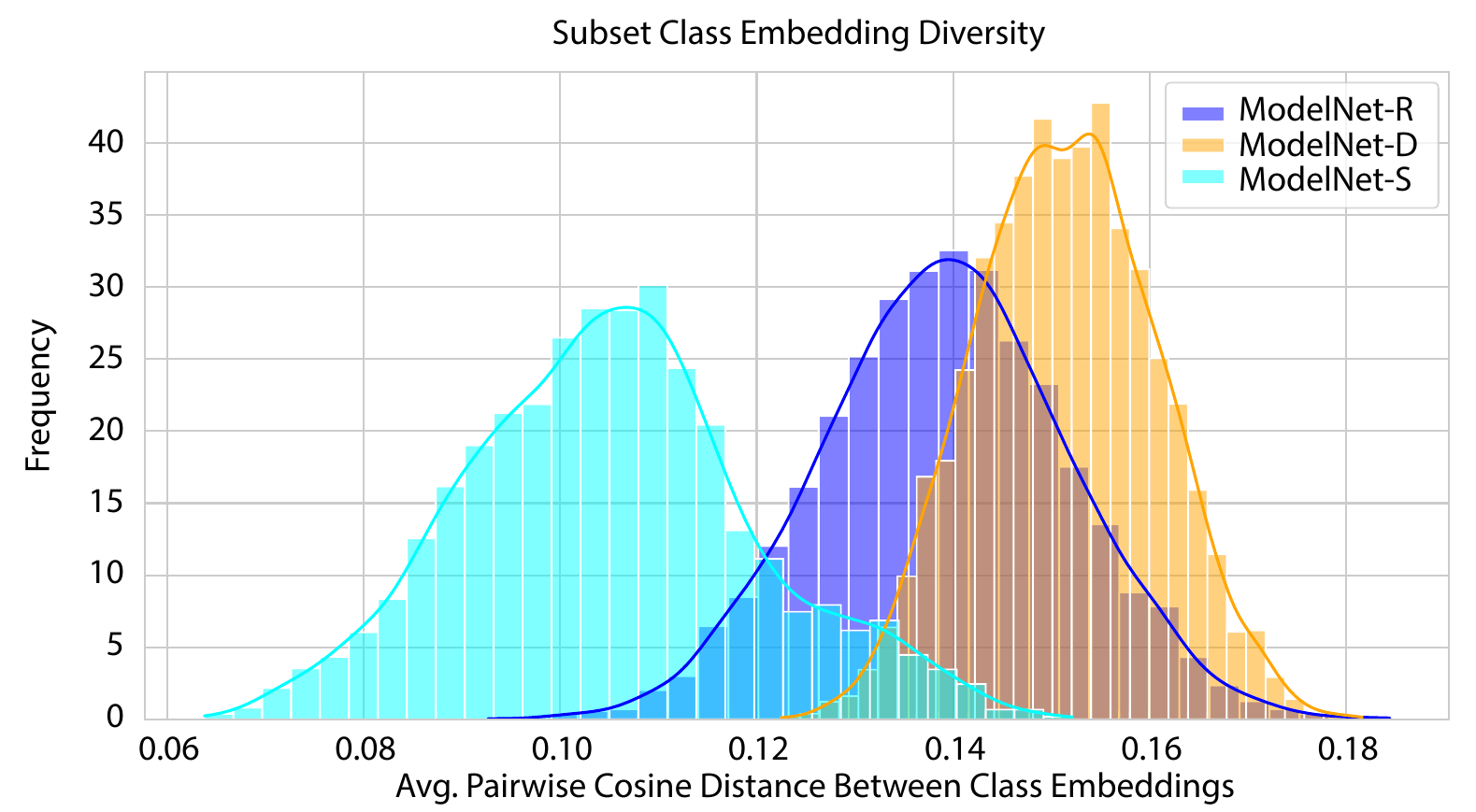}
          \caption{{Subset diversity of ModelNet-R, ModelNet-D, and ModelNet-S.}}
          \label{fig:subsetdiversity}
        \end{figure}

        Fig.~\ref{fig:subsetdiversity}  presents the distribution of $\mathcal{D}(S)$ values across 5000 subsets for each of the ModelNet variants. Each histogram is overlaid with a kernel density estimate (KDE) to facilitate comparative analysis.
        ModelNet-D exhibits the highest mean diversity, followed by ModelNet-R and ModelNet-S, with average cosine distances peaking around 0.15, 0.14 and 0.10, respectively. The distribution of ModelNet-D is unimodal and slightly skewed right, indicating that most subsets exhibit a consistently high degree of semantic dissimilarity. 
        The histogram of ModelNet-R is symmetric and narrower, suggesting greater consistency in subset construction. 
        The histogram of ModelNet-S, in contrast, shows a pronounced left shift with a peak near 0.10 and a broader spread. This indicates that a significant portion of its subsets have tightly clustered class embeddings, reflecting lower semantic diversity. 
        \begin{table*}[h!]
            \begin{center}
                \tabcolsep=0.13cm
                \centering
                \captionsetup{justification=centering}
                \caption{
                {Subset Diversity Across ModelNet Variants}}
                    \begin{tabular}{ccccc}
                        \textbf{Dataset} & \textbf{Peak Dist.} & \textbf{Shape} & \textbf{Diversity Level} & \textbf{Implication} \\ \hline
                        ModelNet-D & \(\sim 0.15\) & Right-skewed & High & High generalization, diverse subsets \\
                        ModelNet-R & \(\sim 0.14\) & Symmetric & Moderate & Balanced and consistent subsets \\
                        ModelNet-S & \(\sim 0.10\) & Left-skewed & Low & Fine-grained, semantically tight subsets \\
                        \bottomrule
                        \bottomrule
                    \end{tabular}
                \label{table:subset_diversity}
            \end{center}
        \end{table*}
        
        \textbf{Implications:} These findings underscore the utility of embedding-based diversity metrics in guiding dataset selection and subset generation strategies for representation learning tasks. Depending on the desired trade-off between generalization and specialization as shown in Table \ref{table:subset_diversity}, one can leverage ModelNet-D for exploration of diverse visual semantics, or ModelNet-S for tasks emphasizing intra-class similarity and compact decision boundaries.

    \paragraph{\textbf{Class Occurrence Histogram}}
        \begin{figure*}
            \centering
            \includegraphics[width=.95\linewidth]{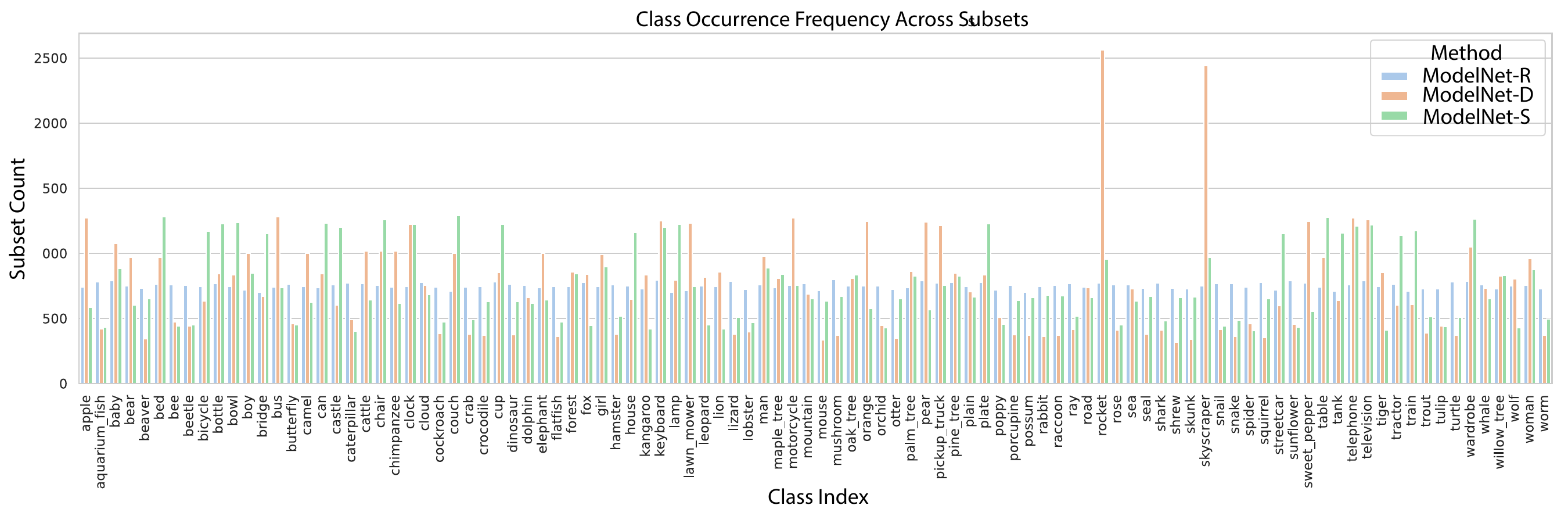}
          \caption{{Class occurrence histogram  of ModelNet-R, ModelNet-D, and ModelNet-S.}}
          \label{fig:class-histo}
        \end{figure*}
        Fig.~\ref{fig:class-histo} presents a histogram of class occurrence frequencies across subsets for three variants of the ModelNet dataset
        The x-axis corresponds to class labels in CIFAR-100, and the y-axis shows the frequency with which each class appears across generated subsets.
        The class occurrence distribution in ModelNet-R is comparatively uniform across all classes, with most bars hovering around a similar count (700). 
        This baseline serves as a reference for interpreting the deviations observed in the other two methods.
        The ModelNet-D variant demonstrates marked variance in class frequency, with certain classes 
        appearing in a significantly larger number of subsets---more than 2500 times in the most extreme case. This suggests that the diversity-based sampling strategy repeatedly selects these classes, 
        thereby maximizing inter-class dissimilarity. 
        Conversely, ModelNet-S, which emphasizes intra-subset similarity, exhibits higher frequencies for a different set of classes.
        These classes may lie in dense regions of the feature space, making them frequent candidates for similarity-driven grouping. The frequency spread in ModelNet-S is narrower than in ModelNet-D but still reveals noticeable preference for certain classes. This bias could result from tight semantic or visual clusters formed during class embedding, reflecting the tendency of the sampling procedure to over-represent such clusters.
        \newline 
        \textbf{Implications:} 
        The observed differences in class frequency distribution across the three methods indicate that the choice of sampling strategy has substantial implications for downstream evaluation. Diverse selection enhances coverage of rare classes but risks over-representation of outliers, while similarity-based sampling fosters coherent intra-subset semantics but may under-represent broader class diversity. Thus, benchmarking models on these subsets without controlling for class distribution may confound performance with dataset construction biases.

        \begin{figure*}
          \centering
            \centering
            \includegraphics[width=\linewidth]{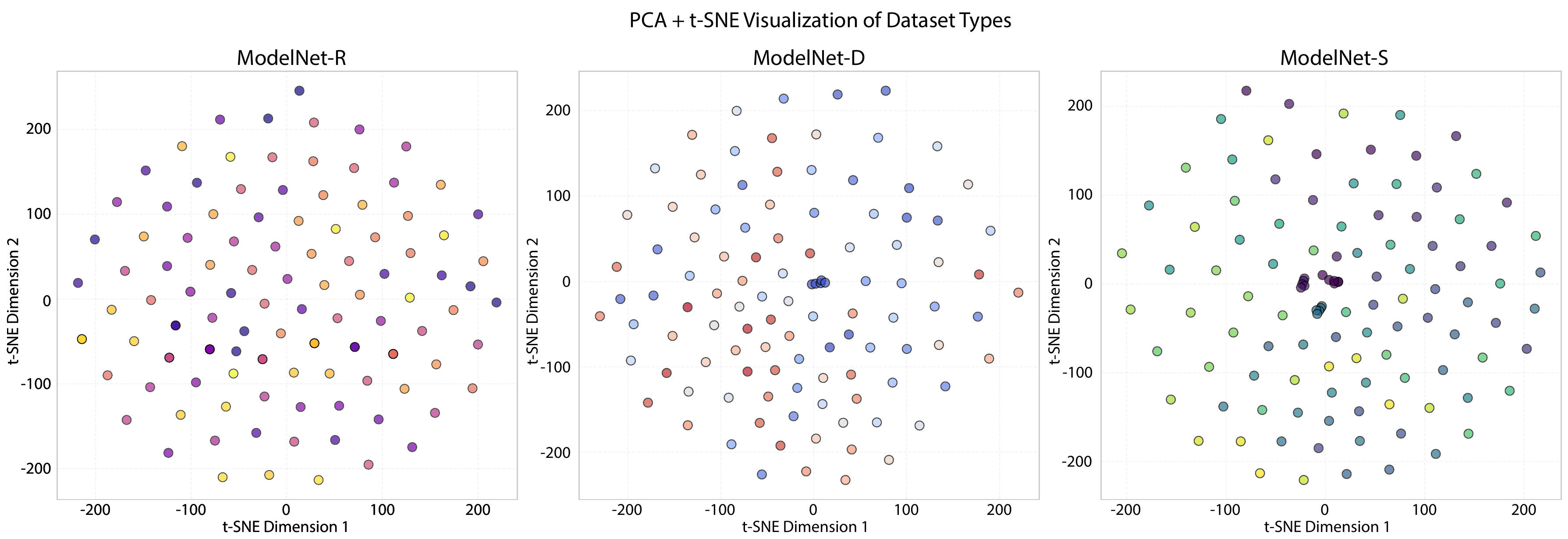}
          \caption{{Three t-SNE plot of ModelNet-R, ModelNet-D, and ModelNet-S.}}
          \label{fig:t-sne}
        \end{figure*}


    \paragraph{\textbf{t-SNE Visualization of Dataset Selection Strategies}}
        Fig.~\ref{fig:t-sne} present a qualitative analysis of the distributional characteristics of datasets generated using different subset selection strategies. To enable visualization, high-dimensional feature representations—extracted from a pretrained deep model—are first reduced via Principal Component Analysis (PCA) and subsequently embedded into two dimensions using t-distributed Stochastic Neighbor Embedding (t-SNE) \cite{van2008visualizing}. Each subplot illustrates the resulting 2D projections, where individual points correspond to class-level embeddings and are color-coded according to their respective class labels.
        %
        
        
        
        
        The ModelNet-D subset exhibits a relatively even and widespread distribution across the embedding space. The visible separation between clusters suggests that the ModelNet-D strategy effectively captures diverse and semantically distinct classes, aligning with its objective of maximizing class embedding diversity. 
        In contrast, the ModelNet-S subset shows tighter clustering with several overlapping regions. This indicates that selected subsets tend to be locally concentrated in the feature space, emphasizing intra-cluster similarity as desired to emphasize local structure. 
        The random selection in the ModelNet-R dataset yields a broader spread than ModelNet-S but lacks the structured dispersion of ModelNet-D. While random sampling does cover various regions, it does so without prioritizing semantic separation or representational balance, often resulting in redundant or overlapping classes.
        \newline 
        \textbf{Implications:} These results highlight how diversity-aware selection (ModelNet-D) can produce more balanced and distinguishable representations, which may lead to improved model generalization and coverage of the feature space.

    \paragraph{\textbf{Jaccard Similarity for Cross-Variant Subset Overlap Assessment}}

        \begin{figure}
          \centering
          \captionsetup{justification=centering}
            \includegraphics[width=\linewidth]{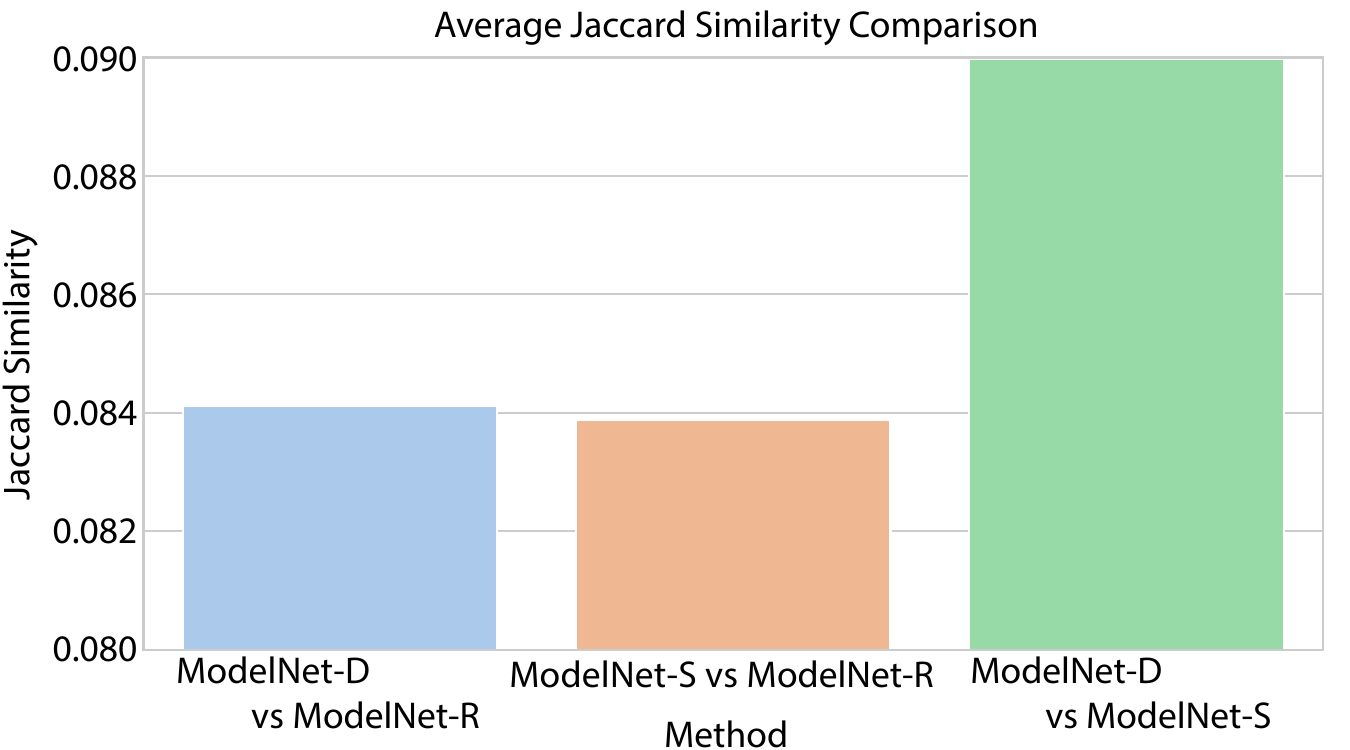}
          \caption{{Jaccard similarity of ModelNet-R, ModelNet-D, and ModelNet-S.}}
          \label{fig:jaccard}
        \end{figure}
    
        We compute the average pairwise Jaccard similarity between the generated subsets.
        Fig.~\ref{fig:jaccard} illustrates the comparative average Jaccard similarities across all subset pairs.

        
        The observed Jaccard similarities for ModelNet-D vs ModelNet-R ($\sim0.084$) and ModelNet-S vs ModelNet-R ($\sim0.0839$) are relatively low, reflecting the distinct nature of deterministic selection strategies compared to random sampling. This indicates that both diversity- and similarity-based methods consistently select class combinations that diverge from the distributions generated by purely random procedures. The near-equivalence of the scores suggests that the degree of departure from randomness is comparably strong in both structured approaches.
        %
        Surprisingly, ModelNet-D vs ModelNet-S exhibits a noticeably higher average Jaccard similarity ($\sim0.09$). Despite the opposing selection objectives—maximizing dissimilarity versus enforcing similarity—this result suggests that both strategies converge on a common subset of frequently selected classes. This convergence is likely driven by the underlying class embedding structure, where certain class groups 
        simultaneously satisfy both diversity and similarity criteria. These classes may inhabit sparsely populated regions of the class embedding space or form tight sub-clusters that align with both optimization heuristics.
        \newline 
        \textbf{Implications:} These results underscore the importance of analyzing inter-subset similarity when constructing evaluation benchmarks. While low overlap ensures broad coverage and reduced redundancy, the elevated similarity between ModelNet-D and ModelNet-S suggests that selection biases—particularly toward prominent or structurally salient classes—can manifest across strategies. Consequently, model evaluations across these variants may not be fully independent, and care should be taken to account for underlying class frequency imbalances.

    \paragraph{\textbf{Subset Redundancy}}
    
        \begin{figure}
          \centering
            \centering
            \includegraphics[width=\linewidth]{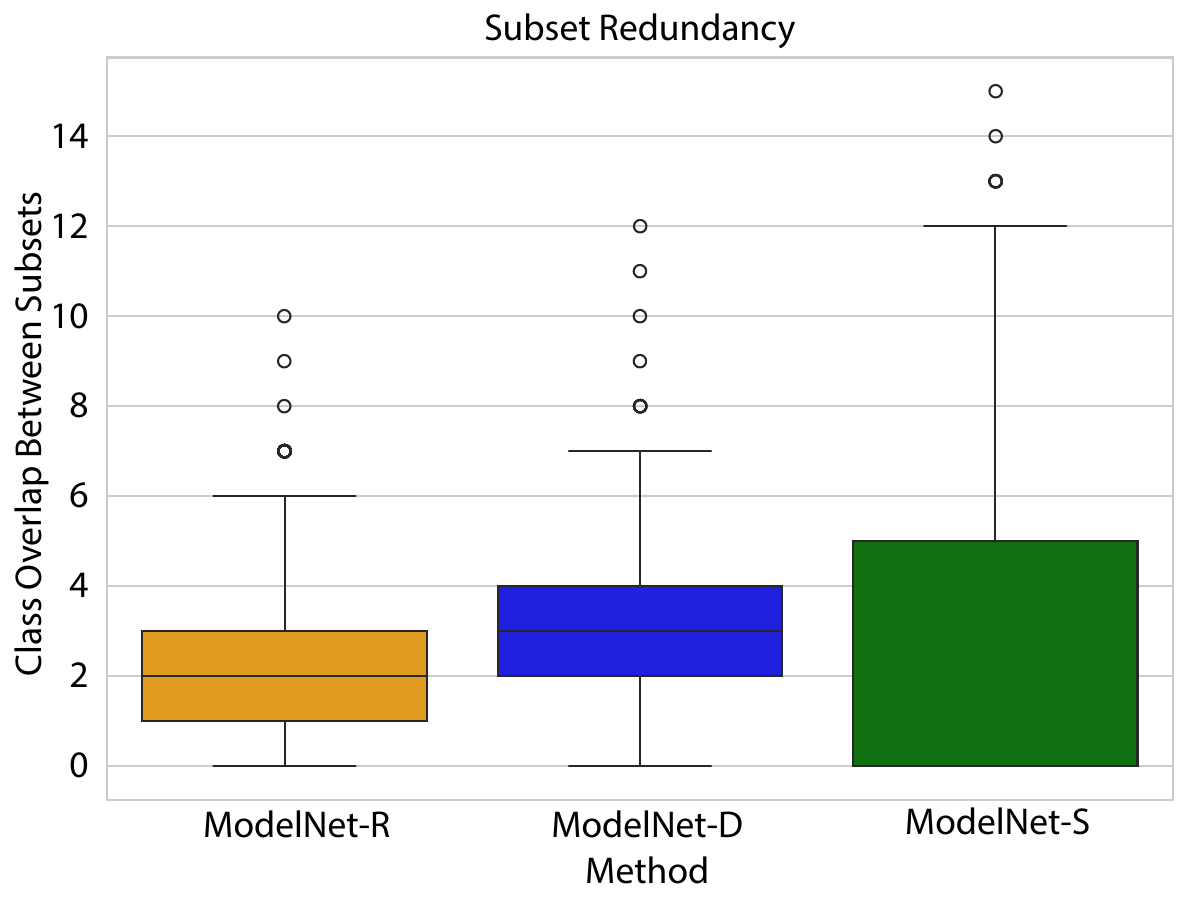}
          \caption{{Subset redundancy of the ModelNet datasets.}}
          \label{fig:subset_redundancy}
        \end{figure}

        To characterize the overlap and redundancy of classes among generated subsets, we perform an in-depth analysis of class co-occurrence across subsets. Redundancy here serves as a proxy for the distinctiveness of subsets and has direct implications for the robustness and fairness of downstream evaluation tasks. 
        Fig.~\ref{fig:subset_redundancy} illustrates the distribution of class overlaps using a boxplot representation across all subset pairs within each sampling strategy. ModelNet-R demonstrates the lowest redundancy, with a median class overlap near 2 and minimal upper-bound outliers. The randomness of class selection ensures a high degree of heterogeneity across subsets. This results in broader coverage of the class space and enhances the generalizability of evaluations.
        Surprisingly, ModelNet-D exhibits slightly higher redundancy than the random baseline. Although designed to maximize diversity in embedding space, it inadvertently concentrates on classes that are semantically or structurally distinct—often leading to repeated selections across subsets. This is reflected in both an increased median overlap and a greater number of high-outlier cases, including overlaps exceeding 10 classes.
        As expected, ModelNet-S results in the highest redundancy. Subsets generated under this regime often draw from densely clustered regions in the class embedding space, promoting frequent reuse of similar class labels. The median overlap increases significantly, with outliers reaching up to 15 shared classes between subsets—indicating poor separation among them.
        \newline 
        \textbf{Implications:} These findings underscore an essential trade-off between semantic control in class selection and global diversity across subsets. While methods like ModelNet-D and ModelNet-S serve distinct experimental goals (e.g., robustness to inter-class similarity or diversity), their increased subset redundancy can bias evaluation outcomes by inflating familiarity across test splits.

    \paragraph{\textbf{Intra-Subset Variance}}
        \begin{figure}
            \centering
            \captionsetup{justification=centering}
            \includegraphics[width=\linewidth]{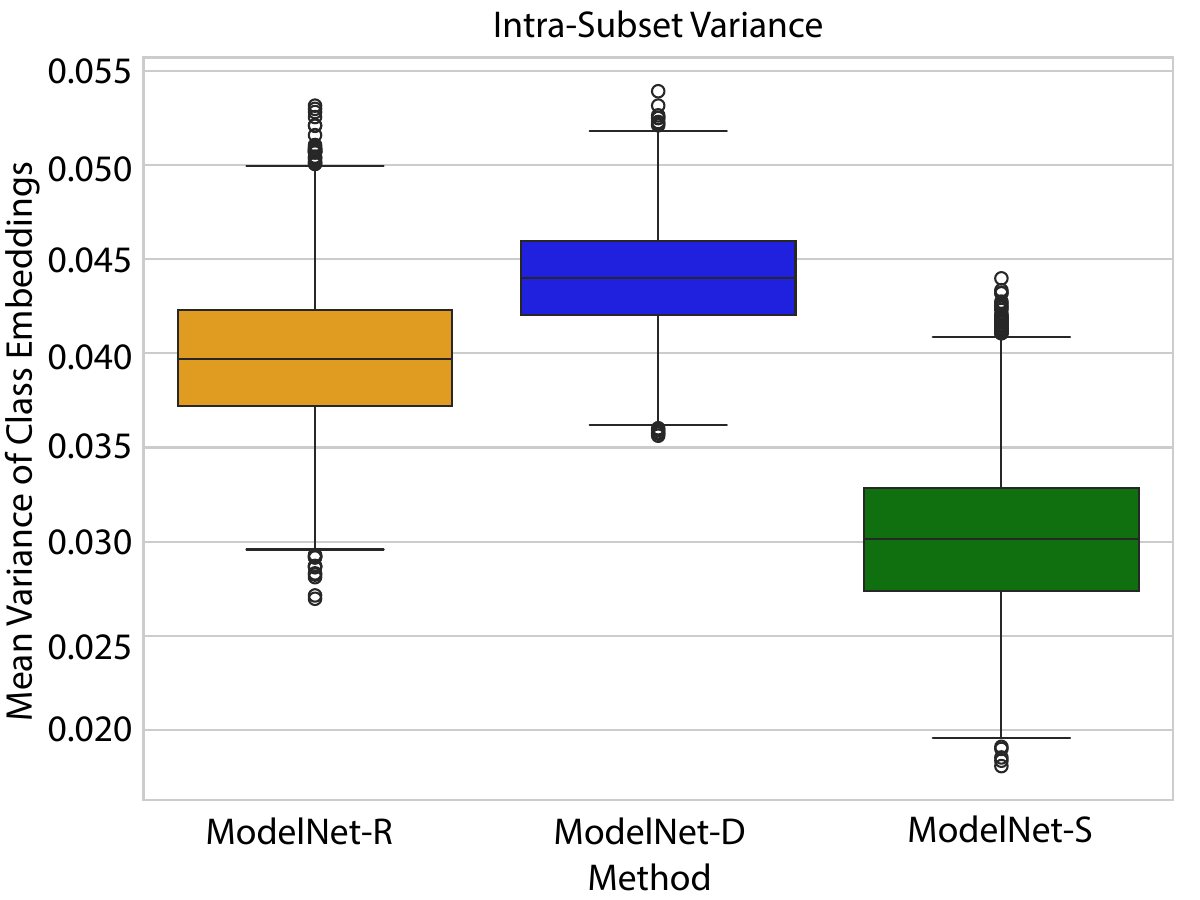}
          \caption{{Intra-subset variance of the ModelNet datasets.}}
          \label{fig:intra-subset}
        \end{figure}
    
        Fig.~\ref{fig:intra-subset} presents the distribution of intra-subset variance for ModelNet-R, ModelNet-D, and ModelNet-S across multiple subsets. ModelNet-D exhibits consistently higher intra-subset variance that indicates the underlying KNN-based selection strategy results in class embeddings that are more dispersed and diverse. In contrast, ModelNet-S shows the lowest variance, suggesting a more compact feature space per class—likely due to its semantic proximity constraint. ModelNet-R falls in between, balancing diversity and compactness as class labels are randomly selected for each subset. 
        \newline 
        \textbf{Implications:} These observations confirm that different subset selection strategies induce distinct intra-class spread characteristics in the feature space, which can have important implications for downstream multi-domain FL tasks.

    \paragraph{\textbf{Feature Space Coverage}}
       \begin{figure}
            \centering
            \captionsetup{justification=centering}
            \includegraphics[width=\linewidth]{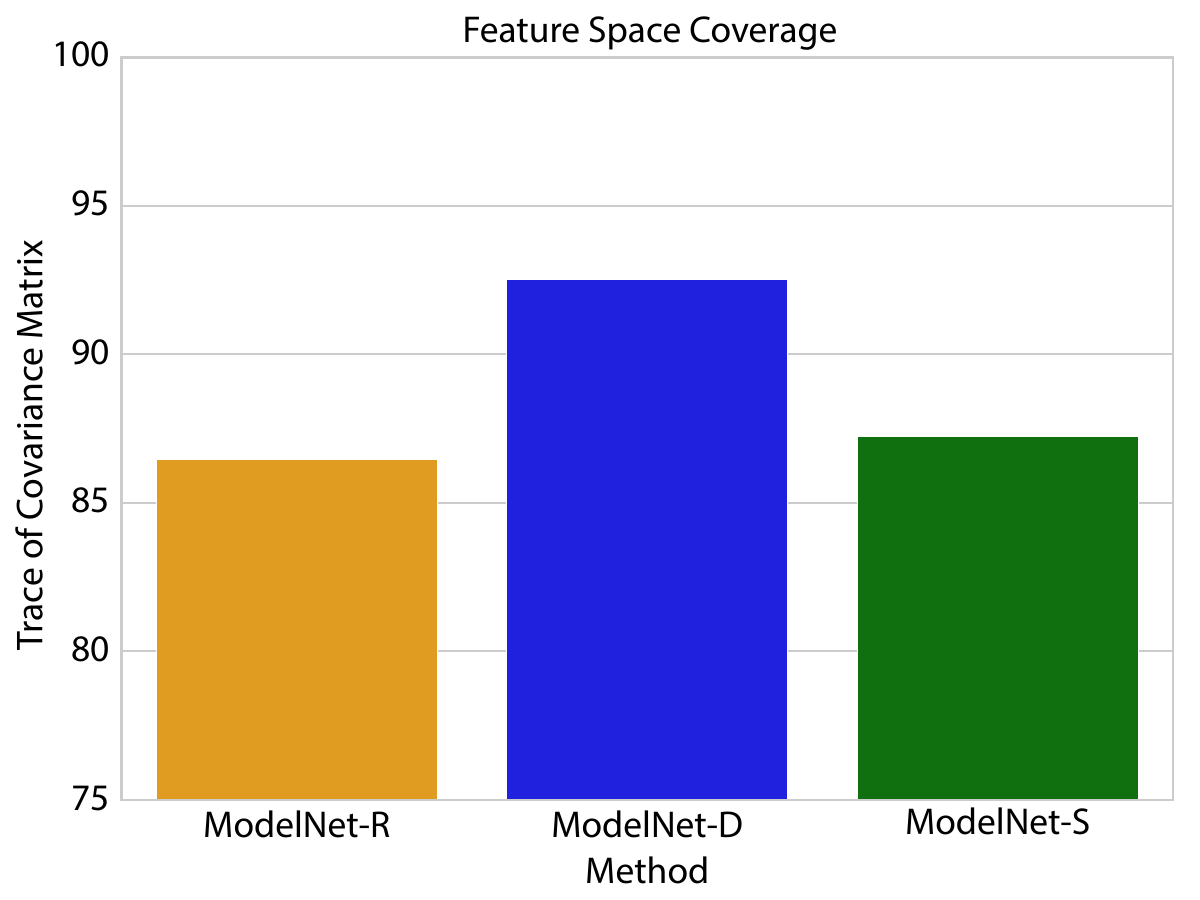}
          \caption{{Feature space coverage of the ModelNet datasets.}}
          \label{fig:feature-space-coverage}
        \end{figure}
        
       Fig.~\ref{fig:feature-space-coverage} reports the feature space coverage for the ModelNet datasets. Among the three, ModelNet-D achieves the highest trace value, indicating a broader and more diverse utilization of the embedding space. This suggests that the distance-based selection strategy used in ModelNet-D promotes feature spread and maximizes coverage. In contrast, ModelNet-S, which is guided by semantic similarity, results in the lowest coverage, implying that its feature representations are more concentrated and localized. ModelNet-R shows moderate coverage, serving as a baseline.
        \newline 
        \textbf{Implications:} These results highlight the impact of subset construction strategies on the representational diversity of the dataset, which can influence model generalization and transferability.
        




\section{Conclusions}
\label{sec:Conclusions}
In this work, we introduce ModelNet, a large-scale and versatile federated learning (FL) benchmark specifically designed to address the limitations of existing datasets in modeling statistical and semantic heterogeneity across clients. We enable systematic control over class distribution and semantic similarity by constructing three distinct variants: ModelNet-R, ModelNet-D, and ModelNet-S. It offers a novel framework for evaluating FL algorithms under realistic and diverse conditions. Beyond traditional data-based benchmarking, ModelNet uniquely supports graph-based model interpretation, laying the groundwork for privacy-preserving and structure-aware FL research. Our anonymized parameter-sharing approach for dataset construction enhances privacy while maintaining model utility. The dataset is readily extensible to other base image classification datasets beyond CIFAR100 and architectures beyond ResNet50 due to adaptive underlying data distribution algorithms. Extensive experiments validate the effectiveness of each variant of ModelNet for inter- and intra-client data distribution environments. By open-sourcing the dataset and evaluation tools, we aim to establish ModelNet as a comprehensive, reproducible, and future-ready benchmark for both classical and graph-driven FL research.

\section{Acknowledgments} This work was supported by the FLOCKD project funded by the 
DFF
under the grant agreement number 1032-00179B.

\bibliographystyle{ACM-Reference-Format}
\bibliography{sample-base}

\end{document}